\documentclass{article}
\usepackage{ijcai21}

\usepackage{times}
\usepackage{soul}
\usepackage{url}
\usepackage[hidelinks]{hyperref}
\usepackage[utf8]{inputenc}
\usepackage[small]{caption}
\usepackage{graphicx}
\usepackage{amsmath,amssymb}
\usepackage{amsthm}
\usepackage{booktabs}
\usepackage{algorithmic}
\usepackage{latexsym}

\usepackage{enumerate}
\usepackage{subfigure}
\usepackage{csquotes}
\usepackage{xcolor}
\usepackage[ruled,vlined]{algorithm2e}

\SetKwInput{KwData}{Input}
\SetKwInput{KwResult}{Output}

\title{Incorporating Domain Knowledge into Deep Neural Networks}

\author{
	Tirtharaj Dash$^{1,3}$, Sharad Chitlangia$^{1,2}$, Aditya Ahuja$^{1,3}$, Ashwin Srinivasan$^{1,3}$
	\affiliations
	$^1$APP Centre for Artificial Intelligence Research (APPCAIR) \\
	$^2$Department of Electrical and Electronics Engineering \\
	$^3$Department of Computer Science and Information Systems \\
	BITS Pilani, K.K. Birla Goa Campus, Goa 403726, India
	\emails
	\{tirtharaj,f20170472,f20170080,ashwin\}@goa.bits-pilani.ac.in
}

\begin{document}
	
	\maketitle
	
	\begin{abstract}
		We present a survey of ways in which 
		domain-knowledge has been included when
		constructing models with neural networks.
		The inclusion of domain-knowledge is of
		special interest not just to
		constructing scientific assistants, but
		also, many other areas that involve understanding data
		using human-machine collaboration. In many
		such instances, machine-based model construction
		may benefit significantly from being provided with
		human-knowledge of the domain encoded in a sufficiently precise
		form. This paper examines two broad approaches to encode such
		knowledge--as logical and numerical constraints--and describes
		techniques and results obtained in several sub-categories under
		each of these approaches.
	\end{abstract}
	
	\section{Introduction}
	\label{sec:intro}
	
\begin{figure*}[!htb]
		\centering
		\subfigure[]{\includegraphics[width=0.54\textwidth]{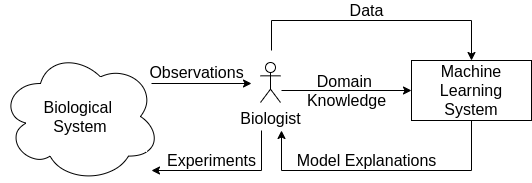}\label{fig:ml_a}} \hfill
		\subfigure[]{\includegraphics[width=0.4\textwidth]{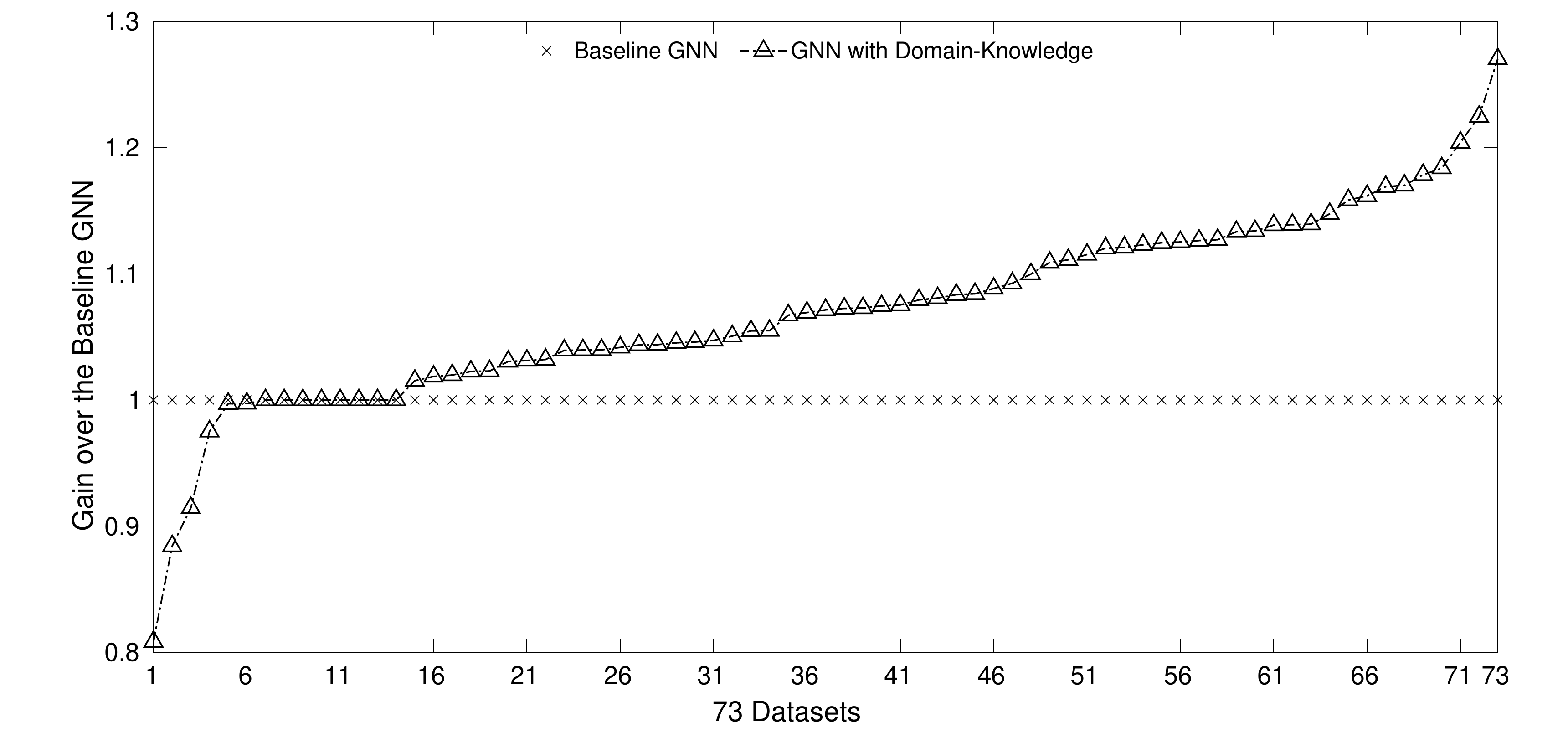}\label{fig:ml_b}}%
		\caption{An example of AI for Science. In (a), the human-in-the-loop
			is a biologist. The plot in (b) from~\protect\cite{dash2020incorporating},
			which shows gains in predictive accuracy of graph neural
			network (GNN) with the inclusion
			of domain-knowledge, using a simplified technique called
			vertex-enrichment. The results shown are over 70 datasets.}
	\end{figure*}

	Artificial Intelligence for Science~\cite{stevens2020ai} is concerned
	with the use of AI methods
	to accelerate our understanding of the natural world,
	and to assist the application of this understanding to
	the development of areas of engineering, medicine, healthcare,
	agriculture, environment and so on. While ambitious plans
	exist for completely automated AI-based robot scientists~\cite{kitano2016artificial},
	the main use of AI for Science remains semi-automated, with
	a scientist-in-the-loop. An example of such a collaborative
	system is in Fig.~\ref{fig:ml_a}. For such systems
	to work effectively, we need at least the following:
	(1) We have to be able to tell the machine what we know,
	in a suitably precise form; and (2) The machine has to
	be able to tell us what it is has found, in a suitably
	understandable form. While the remarkable recent successes
	of deep neural networks on a wide variety of tasks makes
	a substantial case for their use in model construction, it
	is not immediately obvious how either (1) or (2) should be
	done with deep neural networks. In this paper, we examine
	ways of achieving (1). Understanding models constructed
	by deep neural networks is an area of intense research
	activity, and good summaries exist elsewhere \cite{lipton2016he,arrieta2019xplainable}.
	To motivate the utility of providing domain-knowledge to
	a deep network, we reproduce a result from \cite{dash2020incorporating}
	in Fig.~\ref{fig:ml_b}, which shows that
	predictive performance can
	increase significantly, even with a simplified encoding
	of domain-knowledge.
	It is unsurprising that a recent report on
	AI for Science~\cite{stevens2020ai} identifies
	the incorporation of domain-knowledge as one of the 3
	Grand Challenges in developing AI systems:
	
	\begin{displayquote}
		``Off-the-shelf [ML and AI]  practice
		treats [each of these]
		datasets in the same way and ignores domain knowledge
		that extends far beyond the raw data\ldots
		Improving our ability to systematically incorporate
		diverse forms of domain knowledge can
		impact every aspect of AI.''
	\end{displayquote}
	
	In this survey, we restrict the studies on incorporation
	of domain knowledge into neural networks, with 1 or 
	more hidden layers (we will sometimes also use the term deep
	neural network, or DNN). Before we proceed further, we
	clarify that our focus here are more specific than
	that of research that looks at the development of
	hybrid systems combining neural
	and logical systems (see
	for example, ~\cite{garcez2012neural}); and different
	to the use of neural network techniques for either emulating logical
	inference or to represent
	logical concepts. We 
	refer the reader to \cite{besold2017neural}
	for reviews of some of these other strands of work. These reviews
	are nevertheless relevant to some of the material
	in this paper since they identify some key challenges in 
	integrating neural-based learning with
	symbolic knowledge representation and logical reasoning.
	More directly related to this paper is the work
	on ``informed machine learning'', reviewed in ~\cite{von2019informed}.
	We share with this work the interest in prior knowledge as an
	important source of information that can augment existing data.
	However, the goals of that paper are more ambitious than here. It aims
	identify categories of prior knowledge, using as dimensions:
	the source of the knowledge, its representation, and its point of use in
	a machine-learning algorithm. In this paper, we are only concerned
	with some of these categories. Specifically, in
	terms of the categories in \cite{von2019informed}, we are interested
	in implicit or explicit sources of domain-knowledge, represented either
	as logical or numeric constraints, and used at the model-construction
	stage by DNNs.
	
	\subsubsection{Our contributions}
	In this survey, we consider
	the representation of domain knowledge
	for DNNs in two broad categories:
	(1) as logical constraints; and
	(2) as numerical constraints.\footnote{We use the term
		``constraints'' here in the sense that the
		domain-knowledge constrains either the structure or parameters
		(or both) of a DNN.}
	Under each of these, we consider sub-categories:
	(1a) Propositional logic, including canonical normal forms;
	(1b) Predicate logic, including binary relations, and
	more generally, $n$-ary relations; 
	(2a) Loss functions, including regularisation terms;
	(2b) Constraints on weights, including priors and transfer-learning.
	
	The rest of the paper is organised as follows:
	Section~\ref{sec:logic} describes inclusion of domain-knowledge
	as logical constraints. Section \ref{sec:numeric} describes
	the inclusion of domain-knowledge as numeric constraints.
	Section~\ref{sec:challenge} outlines some major challenges
	related to the inclusion of domain-knowledge in DNNs
	with our perspectives on on the relevance of the use of domain-knowledge to aspects of Responsible AI, including
	ethics, fairness, and explainability of DNNs.

	\section{Domain-Knowledge as Logical Constraints}
	\label{sec:logic}
	
	\subsection{Propositional Logic}
	
	Although not all DNNs require data to be
	a set of feature-vectors, this form of data
	representation is long-standing and still
	sufficiently prevalent. In logical terms,
	we categorise feature-based representations
	as being encodings in a propositional logic. The
	reader would point out, correctly, that feature-values
	may not be Boolean. This is correct, but we can represent
	non-Boolean features by Boolean-valued propositions (for
	example, a real-valued feature $f$ with value $4.2$ would
	be represented by a corresponding Boolean feature $f'$ that
	has the value $1$ if $f=4.2$ and $0$ otherwise). With the
	caveat of this rephrasing, it has of course been possible
	to provide domain-knowledge to neural networks by employing
	domain-specific features devised by an expert.
	However, we focus here on ways in which domain-knowledge encoded as rules in
	propositional logic has been used to constrain
	the structure or parameters of models constructed by a network.
	
	
	Here, domain-knowledge is encoded as a set
	of propositional rules. These rules in turn 
	constrain the structure of the neural network.
	Weight-learning then proceeds as normal, using
	the structure. The result could be thought of
	as learning weighted forms of the antecedents
	present in the rules.
	The most popular and oldest work along this line
	is Knowledge-Based Artificial Neural Network (KBANN)~\cite{towell1990refinement}
	that incorporates knowledge into neural networks. 
	In KBANN, the domain knowledge is 
	represented as a set of hierarchically structured propositional rules that directly determines a
	fixed topological structure of a neural network~\cite{towell1994knowledge}. 
	KBANN was successful in many real-world applications; but,
	its representational power was bounded by
	pre-existing set of rules which restricted it to
	refine these existing rules rather than discovering new
	rules. A similar study is KBCNN~\cite{fu1993knowledge},
	which first identifies and links
	domain attributes and concepts consistent with
	initial domain knowledge.
	Further, KBCNN introduces additional hidden
	units into the network and most importantly,
	it allowed decoding of the learned rules from
	the network in symbolic form. However, both
	KBANN and KBCNN were not appropriate for learning
	new rules because of the way the initial structure
	was constructed using the initial domain knowledge base.
	
	Some of the limitations described above could be
	overcome with the proposal of a hybrid system
	by Fletcher and Obradovic~\cite{fletcher1993combining}.
	The system was able to learn a neural
	network structure that could construct new rules from
	an initial set of rules. Here, the domain knowledge is
	transformed into an initial network through an
	extended version of KBANN's symbolic knowledge encoding.
	It performed incremental hidden unit
	generation thereby allowing construction or
	extension of initial rule-base. In a similar manner,
	there was a proposal for using Cascade ARTMAP~\cite{Tan1997} which could not only 
	construct a neural network structure from 
	rules but also perform explicit cascading of rules
	and multistep inferencing. It was found that 
	the rules extracted from Cascade ARTMAP
	are more accurate and much cleaner than the rules
	extracted from KBANN~\cite{towell1993extracting}.
	
	In the late 1990s, Garcez and Zaverucha proposed
	a massively parallel computational model
	called CIL$^2$P based on
	feedforward neural network that integrates
	inductive learning from examples and domain knowledge,
	expressed as a propositional logic program~\cite{AvilaGarcez1999}.
	A translation algorithm generates a neural network.
	Unlike KBANN, the approach uses the notion of 
	``bipolar semi-linear'' neurons. This allows the
	proof of a form of correctness, showing the existence of
	a neural-network structure that can compute the logical
	consequences of the domain-knowledge.
	The output of such a network, when combined into
	subsequent processing naturally incorporates the intended
	interpretation of the domain predicates.
	The authors extend this to the use of first-order logic
	programs: we consider this in a later section.
	
	
	A recent proposal focuses on embedding symbolic
	knowledge expressed as logical rules~\cite{Xie2019}.
	It considers two languages of representations:
	Conjuctive Normal Form (CNF) and
	decision-Deterministic Decomposable Negation
	Normal form (d-DNNF), which can naturally 
	be represented as graph structures. The graph
	structures can be provided to a graph neural network (GNN) to
	learn an embedding suitable for further
	task-specific implementations. 
	
	Somewhat in a similar vein to the work
	by \cite{AvilaGarcez1999}, the work reported
	in ~\cite{Xu2018} considers
	as a set of propositional statements representing
	domain constraints. A deep
	network is then trained to find satisfying assignments
	for the constraints. Again, once such a network is constructed,
	it can clearly be used in subsequent processing, capturing
	the effect of the domain constraints. The network is
	trained using a semantic loss that we describe in a
	later section.
	
	\subsection{First-Order Logic}
	
	We now describe methods that use domain-knowledge
	described in first-order logic, which clearly gives
	greater flexibility and representational power than
	the use of propositional logic. Although there
	is no conceptual need to do so, we nevertheless first consider the encoding of domain-knowledge
	as binary relations, and later consider the
	general case. The reasons for this will be apparent
	below.
	
	\subsubsection{Binary Relations}
	
	An influential form of representing domain-knowledge
	takes the form {\em knowledge graph\/}, which
	are labelled graphs, with vertices representing
	entities and edges representing relations between
	entities. In essence, this represents a binary
	relation. We refer the reader to \cite{hogan2020knowledge} to
	a comprehensive survey of this form of representation
	for domain-knowledge.
	
	Incorporation of the information in a knowledge-graph
	into deep neural models--termed ``knowledge-infused
	learning''--is described in \cite{kursuncu2019nowledge,sheth2019shades}.
	This aims to incorporate binary relations contained
	in application-independent sources (like DBPedia, Yago, WikiData)
	and application-specific sources (like SNOMED-CT, DataMed).
	The work examines techniques for incorporating relations
	at various layers of deep-networks (the authors categorise
	these as ``shallow'', ``semi-deep'' and ``deep'' infusion). In the case of shallow infusion,
	both the external knowledge and the method
	of knowledge infusion is shallow, utilising syntactic and lexical
	knowledge in the form of word embedding models. In semi-deep infusion, external knowledge is involved through attention mechanisms or learnable knowledge constraints acting as a sentinel
	to guide model learning, and deep infusion employs a stratified
	representation of knowledge representing different levels of abstractions in different layers of a deep learning model, to transfer
	knowledge that aligns with the corresponding layer in the layered learning process.
	
	Knowledge graphs can be encoded directly
	for use by a graph neural network (GNN). The computational
	machinery available in GNN then aggregates
	and combines the information available in the 
	knowledge graph. The final collected information
	from this computation could be used for further
	predictions. Some recent works are in~\cite{10.1145/3292500.3330855,10.1145/3308558.3313417},
	where a GNN is used for estimation of
	node importance in the knowledge-graph. The idea of
	encoding a knowledge graph directly for a GNN
	is also used in~\cite{chen2019knowledgebased} to enrich
	the information provided in the knowledge-graph.

	\subsubsection{$n$-ary Relations}
	
	The pre-eminent form of
	symbolic machine learning based on
	the use of relations in first-order logic is Inductive Logic Programming (ILP)~\cite{muggleton1991inductive}, which has
	an explicit role for domain-knowledge being incoprated into learning.
	In ILP, domain-knowledge is
	represented in first-order logic: we refer the
	reader to surveys, both old and new, for a description
	of the field \cite{mugg1994ilp,muggleton2012ilp,cropper2020turning}.
	
	The simplest use of ILP to incorporate
	$n$-ary relations in domain knowledge into a neural network
	relies on techniques that automatically ``flatten'' 
	the domain-knowledge into a domain-specific propositional representation. Techniques for automatic construction of
	Boolean-valued features from relational
	domain-knowledge have a long history in the field
	of ILP,  often called {\em propositionalisation\/},
	originating from the LINUS~\cite{lavravc1991learning}.
	This involves the construction of
	features that identify the conditions under which they take on
	the value $1$ (or $0$). For example, given (amongst other
	things) the definition of benzene rings and of fused rings,
	an ILP-based propositionalisation 
	may construct the Boolean-valued feature that has the value $1$ if
	a molecule has 3 fused benzene rings, and $0$ otherwise.
	The values of such Boolean-valued features allows us to
	represent a data instance (like a molecule) as a Boolean-valued
	feature-vector, which can then provided to a neural network.
	There is a long history of
	propositionalisation: see \cite{Kramer2001} for a review of some
	of early use of this technique, and \cite{lavrac2020prop,vig2017investi}
	who examine the links between propositionalisation and
	modern-day use of embeddings in neural-networks.
	
	A direct application of propositionalisation, demonstrating its
	utility for deep networks has been its use in Deep Relational
	Machines ~\cite{Lodhi2013}. A DRM is a deep 
	fully-connected neural network
	with Boolean-valued inputs obtained from propositionalisation
	by an ILP engine. 
	In \cite{dash2018large} Boolean-valued features from an
	ILP engine are sampled 
	from a large space of possible features. The sampling technique
	is refined further in~\cite{dash2019discrete}.
	
	The idea of propositionalisation also
	forms the foundation for a method known as
	`Bottom Clause Propositionalisation (BCP)' to propositionalise
	literals of a most-specific clause, or ``bottom-clause''. Given
	a data instance, the bottom-clause is 
	the most-specific first-order clause that entails the data instance,
	given some domain-knowledge. Loosely speaking, the most-specific
	clause can be thought of ``enriching'' the data instance with
	all domain relations that are true, given the
	data instance. The construction of such most-specific clauses
	and their subsequent use in ILP was introduced in \cite{muggleton1995inverse}.   CILP++~\cite{francca2014fast} uses
	bottom-clauses for data instances to construct feature-vectors for
	neural networks. This is an extension to  CIL$^2$P. Here the neural network
	has recurrent connections.
	
	Propositionalisation has conceptual and practical limitations.
	Conceptually, there is no variable sharing
	between two or more first-order features~\cite{dash2018large}. Practically, the space of possible features can be extremely large: this has meant
	that the feature-selection has usually been done separately from
	the construction of the neural network. There are some recent 
	reports on incorporating first-order logic that do not rely on propositionalisation. 
	In \cite{Li2020} it is proposed to augment a language model that uses a
	deep net architecture with additional statements in first-order logic.
	Thus, given domain-knowledge encoded as first-order relations, connections are introduced into the network, based on the logical constraints enforced by the domain-relations. The approach is related somewhat to the
	work in \cite{DBLP:journals/jair/SourekAZSK18} that does not explicitly consider the incorporation
	of domain-knowledge but does constrain a deep neural network's structure
	based on the relational structure within data instances.
	
	A work that does not employ either
	propositionalisation or network augmentation
	considers a combination of symbolic knowledge
	represented in first-order logic
	with matrix factorization
	techniques~\cite{rocktaschel-etal-2015-injecting}. This
	exploits dependencies between textual patterns to generalise
	to new relations. In another study, drawing on representation of $n$-ary relations as hyperedges,
	the technique of {\em vertex enrichment\/} is proposed
	in~\cite{dash2020incorporating}. This provides a
	simplified way to incorporate symbolic 
	domain-knowledge into
	standard graph neural networks (GNNs)~\cite{dash2020incorporating}. 
	Vertex enrichment incorporates first-order
	background relations as additional features 
	into the features associates with the  nodes of a graph provided
	to a GNN. The results reported in the paper show significant
	improvements in the predictive accuracy of GNNs across a large number
	datasets.
	
	We note that newer areas are emerging that use representations
	for domain-knowledge that go beyond first-order logic relations.
	This includes probabilistic first-order logic, as a way of including
	uncertain domain-knowledge \cite{manhaeve2018eepproblog}. One interesting
	way this is being used is to constrain the training of  ``neural predicates'', which represent probabilistic relations that are implemented
	by neural networks. 
	
	\section{Domain-Knowledge as Numerical Constraints}
	\label{sec:numeric}
	
	We now discuss some forms of numeric constraints
	often used as ways of incorporating domain-knowledge
	into a neural network. Some of the methods are standalone
	methods and some of these methods are
	coupled with methods incorporating logical
	constraints to provide a more robust way of integration of
	domain-knowledge into a network.
	
	\subsection{Utility or Loss Functions}
	
	A fairly standard way of incorporating domain-knowledge
	into a deep network is by introducing additional 
	loss terms into the utility (loss) function that
	the network optimises. This function takes on a general form
	like the following:
	\[
	\mathcal{L}_{total} = \alpha\mathcal{L}_{task} + \beta\mathcal{L}_{in} + \gamma\mathcal{L}_{out} + 
	\lambda\mathcal{L}_{model},
	\]
	where $\mathcal{L}_{total}$ is the
	total loss for the deep network-based
	on which training will be carried out,
	$\mathcal{L}_{task}$ denotes to
	the standard task-specific loss (e.g. cross-entropy for classification, mean-squared-error for regression, and so on),
	$\mathcal{L}_{in}$ is the loss representing some
	constraints on the inputs, and
	$\mathcal{L}_{out}$ is the loss
	representing constraints on outputs, 
	$\mathcal{L}_{model}$ is the constraint
	on the function that is to be learned
	by the deep network. The parameters
	$\alpha, \beta, \gamma, \lambda$ are the
	constants signifying some form
	of weights for each of the above loss terms,
	which can be learned or tuned during model construction.
	
	A recent work that is based on loss function is in~\cite{Xu2018}.
	Here the authors propose a semantic loss that
	signifies how well the outputs of the deep network
	matches some given constraints encoded as
	propositional rules. The general intuition
	behind this idea is that the semantic loss is proportional to a negative logarithm of the probability of generating a state that satisfies the constraint when sampling values according to some probability distribution.
	This loss function falls under the category of
	$\mathcal{L}_{out}$. 
	This kind of loss function is particularly 
	useful for semi-supervised learning as these 
	losses behave like self-information 
	and are not constructed using explicit labels 
	and can thus utilize unlabelled data. 
	
	\cite{Hu2016HarnessingDN} proposed a framework
	to incorporate first-order logic rules with the
	help of an iterative distillation procedure that
	transfers the structured information of logic
	rules into the weights of neural networks. This is
	done via a modification to the knowledge-distillation
	loss proposed by Hinton et al.~\cite{hinton2015istilling}.
	The authors show that taking this loss-based route
	of integrating rule-based domain-knowledge allows
	the flexibility of choosing a deep network architecture
	suitable for the intended task.
	
	In~\cite{Fischer2019DL2TA}, authors construct a system
	for training a neural network with domain-knowledge
	encoded as logical constraints. Here the available
	constraints are transferred to a loss function. Specifically,
	each individual logic operation (such as negation, and, or,
	equality etc.) is translated to a loss term. 
	The final formulation results in an optimisation problem.
	The authors extract constraints on inputs that capture
	certain kinds of convex sets and use them as optimisation
	constraints to make the optimisation tractable. 
	In the developed system, it is also possible to pose queries on the model to find inputs that satisfy a set of constraints.

	\subsection{Constraints on Weights}
	
	In Bayesian framework, explicitly information about a machine
	learning model and data can be expressed succinctly in
	the form:
	\[
	\text{posterior} \propto \text{prior} \cdot \text{sample-likelihood}
	\]
	A very common way of incorporating domain-knowledge into
	a machine learning system (including deep networks) is by
	encoding it as a `prior' term in the above Bayes equation.
	The domain-knowledge here could be about the problem,
	the neural network structure, or the neural network parameters, that is, some form
	of a probability distribution over these.
	The priors on networks and network weights represent
	our expectations about networks before receiving any data, 
	and correspond to penalty terms or regularisers.
	Buntine and Weigend~\cite{Buntine1991BayesianB}
	extensively study how Bayesian theory can be
	highly relevant to the problem of training feedforward
	neural networks. One of the main focus of this study
	was to study principles of choosing an appropriate
	network structure and size based on prior domain-knowledge
	about the problem and also of selecting a prior on the
	weight parameters. 
	
	\subsubsection{Priors}
	
	Seminal work by \cite{hintonnealthesis} on Bayesian 
	learning in neural networks showed how domain-knowledge could help build a prior probability distribution over neural network parameters. 
	They showed how with 
	the approach of Bayesian learning, networks are 
	self-regularised to not over-fit even when complexity of 
	the neural network is increased to infinity. This study has 
	served as foundation to various works on regularisation 
	approaches such as posterior regularisation in the 
	case of neural networks. In a similar spirit,
	\cite{Krupka2007IncorporatingPK} showed how prior domain
	knowledge could be used to define `meta-features'
	that can aid in defining the prior distribution of weights.
	These meta-features are additional information about
	each of the features in the available data. For instance,
	for an image recognition task, the meta-feature could be
	the relative position of a pixel $(x, y)$ in the image. 
	This meta information can be used to construct a prior
	over the weights for the original features.
	
	\subsubsection{Transfer Learning}
	
	Transfer Learning is a very common technique utilised when 
	there is fewer data in the target domain pertaining to 
	the prediction task; and lots of data in the domain of 
	a task similar to the target domain. 
	Transfer learning from a related (source) domain to a 
	target domain results in significant boost in performance
	and saves substantial computational efforts.
	From the Bayesian perspective, transfer learning
	allows the construction of the prior over the weights
	of a neural network for the target domain based on
	the posterior constructed in the source domain. 
	Transfer learning is not limited by the kind of task (such as 
	classification, regression, etc) but rather by the domain 
	itself. This is a result of the multi-layered information 
	learnt by neural networks i.e., the final layer, which 
	is most often corresponding to the task in hand, can be 
	replaced and fine-tuned separately~\cite{YosinskiCBL14}. 
	Large language models
	are a very successful example of this, where the models are
	initially learnt on a huge corpus of data and fine-tuned 
	for numerous tasks. \cite{zhuang2020comprehensive} provides
	an in-depth review of some of the mechanisms and the strategies of transfer learning. Transfer learning need not be
	restricted to deep networks only: in a recent study, 
	\cite{liu2018mproving} proposes a model that transfers
	knowledge from a neural network to a decision tree
	using knowledge distillation framework. The symbolic
	knowledge encoded in the decision tree could further
	be utilised for further for a variety of tasks.
	
	\begin{table*}[!htb]
		\centering
		\begin{tabular}{cccc}
			\toprule
			\textbf{Work}  &  \textbf{Logical Constraint}   &   \textbf{Numerical Constraint}  & \textbf{DNN Method} \\
			\midrule
			KBANN~\cite{towell1994knowledge} & Propositional logic & - & MLP \\
			Cascade-ARTMAP~\cite{Tan1997} & Propositional logic & - & ARTMAP \\
			CIL$^2$P~\cite{AvilaGarcez1999} & Propositional logic & -  & RNN \\
			LENSR~\cite{Xie2019}  & Canonical Normal Form & Loss function & GNN \\
			CILP++~\cite{francca2014fast}   & First-order logic & - & MLP \\
			DRM~\cite{Lodhi2013} & First-order logic & - & MLP \\
			VEGNN~\cite{dash2020incorporating} & First-order logic & - & GNN \\
			Semantic Loss~\cite{Xu2018} & First-order logic & Loss function & CNN\\
			HDNNLR~\cite{Hu2016HarnessingDN} & First-order logic & Loss function & CNN, RNN\\
			DL2~\cite{Fischer2019DL2TA} & - & Loss function & CNN \\
			ILBKRME~\cite{rocktaschel-etal-2015-injecting} & - & Loss function, Regularisation & MLP \\
			IPKFL~\cite{Krupka2007IncorporatingPK} & - & Prior & CNN \\
			\bottomrule
		\end{tabular}
		\caption{Some selected works with category of domain-knowledge and class of DNN method studied}
		\label{tab:summary}
	\end{table*}
	
	\subsection{Regularisation}
	
	A natural consequence of incorporating prior knowledge
	(some knowledge on the prior distribution) results in
	regularisation in learning. This allows the inductive bias
	to be exploited based on prior knowledge and further
	allowing better predictive performance. One of the most common
	regularisations is a penality based regularisation
	for model complexity. Examples include $L_1$ or $L_2$
	based regularisation terms in the utility or loss function
	of a neural network. The optimiser minimises the loss
	along with the regularisation term resulting in a
	less complex model in terms of parameters~\cite{kukavcka2017regularization}. 
	\cite{10.4018/JITR.2018100109} show that domain-based
	regularisation in loss function can also help in constructing deep networks with less amount of data. 
	
	\subsubsection{Regularising Embedding}
	
	Over the years, regularising embedding constitutes another
	major direction of research.
	\cite{Fu1995} was one of the earliest works in this 
	domain, in which they proposed a strategy to 
	establish constraints by designating each node in a
	Hopfield Net to represent a concept and edges to 
	represent their relationships and learn these nets 
	by finding the solution which maximises the greatest 
	number of these constraints. \cite{rocktaschel-etal-2014-low} 
	was perhaps the first method of regularising 
	embeddings from declarative knowledge encoded in 
	first-order logic. \cite{rocktaschel-etal-2015-injecting}
	extended this to regularisation by addition of 
	differentiable loss terms to the objective-based on
	propositionalisation of each first-order predicate.
	\cite{Li2020} develop a method to constraint 
	individual neural layers using soft logic based 
	on massively available declarative rules in ConceptNet. 
	\cite{Hamilton2018EmbeddingLQ} incorporate first-order 
	logic into low dimensional spaces by embedding graphs 
	nodes and represent logical operators as learned 
	geometric relations in the space. 
	\cite{Demeester2016LiftedRI} proposed ordering of 
	embedding space based on rules mined from WordNet 
	and find it to better prior knowledge and 
	generalisation capabilities using these relational embeddings.
	\cite{silvestri2020injecting} probe embeddings of a neural
	network while progressively adding domain knowledge and show 
	strong results encouraging the same.
	In~\cite{takeishi2018knowledge},
	a knowledge-based distant regularisation framework was
	proposed, in which distance domain information encoded in a knowledge graph was utilised. It defines prior distributions of model parameters using knowledge graph embeddings. They show
	that this results in an optimisation
	problem for a regularised factor analysis method.
	
	
	\section{Challenges and Concluding Remarks}
	\label{sec:challenge}
	
	We summarise our discussion on domain-knowledge
	as constraints in Table~\ref{tab:summary}.
	We now outline some challenges in incorporating
	domain-knowledge encoded as logical or numerical
	constraints into a deep network. We first outline
	some immediate practical challenges concerning the
	logical constraints:
	\begin{itemize}
		\item There is no standard framework for translating logical constraints to neural networks.
		While there are simplification methods which
		first construct a representation of the logical
		constraint that a standard
		deep network can consume, this process has its limitations
		as described in the relevant section above.
		\item Logic is not differentiable. This does not allow using standard training of deep network using
		gradient based methods in an end-to-end fashion. 
		Propagating gradients via logic has now been looked
		at in \cite{evans2018delILP}, but the solution is intractable and does not allow day-to-day use.
		\item Neural networks are directed acyclic graphs (DAGs).
		However, logical formula can introduce cyclic dependencies, which needs a separate form of translations.
	\end{itemize}
	There are also practical challenges concerning the numerical constraints:
	\begin{itemize}
		\item We have seen that the numerical constraints are
		often provided with the help of modification to 
		a loss function. Constructing a term in loss function
		is not straight-forward.
		\item The process of introducing a loss term
		often results in a difficult optimisation problem
		(sometimes constrained) to be solved. This may
		require additional mathematical tools for a solution that can be implemented practically.
	\end{itemize}
	
	\subsection*{The Domain-Knowledge Grand Challenge}
	Incorporating domain-knowledge into learning
	is highlighted in \cite{stevens2020ai} as one of the Grand Challenges
	facing the foundations of AI and ML. The principal
	difficulties raised in that report are these:
	\begin{itemize}
		\item ``Can the constructed deep network model be
		trusted?'' This question involves long-standing 
		discussions on explainability and interpretability
		of deep models. It also includes the question of
		whether data used for constructing the deep model 
		contains sufficient information without introducing
		spurious correlations or bias that would invalidate
		the model itself. 
		\item ``Why does the AI model work for a problem?'' 
		To address this question, there has to be some
		a mapping between the internal representation of the model
		to a domain-specific concept.
		In \cite{info11020122}, authors identify that
		the knowledge mapping of the deep learning components,
		including input features, hidden unit and layers,
		and output predictions with domain-knowledge could
		lead to an understandable model.
	\end{itemize}
	
	\subsection*{Going Beyond Prediction}
	
	The issues raised above go beyond just the ``how''
	questions related to the incorporation
	of domain-knowledge into deep networks.
	They provide pointers to why the use of domain-knowledge
	may extend beyond its utility for prediction.
	Domain-knowledge can also
	play a role in aspects like explanation and fairness.
	We mention some of the challenges that result.
	
	One important requirement of machine-constructed models in workflows
	with humans-in-the-loop is that the models are human-understandable. Domain-knowledge can be used in two different ways to assist this. First,
	it can constrain the kinds of models that are deemed understandable.
	Secondly, they can provide concepts that are meaningful for use in
	a model. Most of the work in this review has been focussed on improving
	predictive performance. However, the role of domain-knowledge in constructing
	explanations for deep network models is also being explored
	(see for example, \cite{srinivasan2019logical}). However, that
	work only generates {\em post hoc\/} explanations that are locally
	consistent. Explanatory deep network models that identify true causal connections based on concepts provided as domain-knowledge remain
	elusive.
	
	Domain-knowledge can also be used to correct biases built into
	a deep network either declaratively, through the use of constraints,
	or through the use of loss functions that include``ethical penalty''
	terms. Demonstrations of the use of domain-knowledge driven,
	ethics-sensitive machine learning have been available in
	the literature for some time \cite{Anderson2005MedEthExTA}.
	Can these carry over
	to the construction of deep network models? This remains
	to be investigated.
	
	Finally, the rapid progress in the area of language models
	raises the possibility of providing domain-knowledge in forms
	other than logical or numerical. While the precision of these
	formal representations may continue to be needed for the construction
	of scientific assistants, their role in representing commonsense
	knowledge is less evident. Day-to-day machine assistants 
	that can incorporate informal knowledge of the world will be
	needed. Progress in this is being made (see for example, \url{https://allenai.org/aristo}), but there
	is much more that needs to be done to make the language models
	required accessible to everyday machinery.
	
	\bibliographystyle{named}
	\small{
		\bibliography{ijcai21}
	}

\end{document}